\documentclass[11pt,a4paper]{article}
\usepackage[hyperref]{acl2019}
\usepackage{times}
\usepackage{latexsym}
\usepackage[utf8]{inputenc}
\usepackage{url}

\usepackage[single, list-caps=true]{acro}

\DeclareAcronym{hipaa}{
    short = HIPAA,
    long = Health Insurance Portability Accountability Act,
    extra = {American health care law.},
}
\DeclareAcronym{phi}{
    short = PHI,
    long = protected health information,
}
\DeclareAcronym{ehr}{
    short = EHR,
    long = electronic health record,
}
\DeclareAcronym{nlp}{
    short = NLP,
    long = natural language processing,
    short-indefinite = an,
}
\DeclareAcronym{ner}{
    short = NER,
    long = named entity recognition,
    short-indefinite = an,
}
\DeclareAcronym{crf}{
    short = CRF,
    long = conditional random field,
}
\DeclareAcronym{mlp}{
    short = MLP,
    long = multi-layer perceptron,
    short-indefinite = an,
}
\DeclareAcronym{rnn}{
    short = RNN,
    long = recurrent neural network,
    short-indefinite = an,
}
\DeclareAcronym{lstm}{
    short = LSTM,
    long = long short-term memory,
    short-indefinite = an,
}
\DeclareAcronym{elmo}{
    short = ELMo,
    long = embeddings from language models,
}
\DeclareAcronym{dann}{
    short = DANN,
    long = domain-adversarial neural network,
}

\usepackage{amsmath, amssymb}
\usepackage{bm}
\usepackage{mathtools}
\DeclarePairedDelimiter{\abs}{\lvert}{\rvert}

\usepackage{xspace}
\newcommand{\fone}{\ensuremath{F_1}\xspace}

\usepackage[capitalise]{cleveref}

\usepackage{tikz}
\usetikzlibrary{arrows, backgrounds, calc, scopes, fit, intersections, matrix, positioning, shapes, shapes.misc, shadows}
\usepackage{fontawesome}

\usepackage{framed}

\usepackage[normalem]{ulem}

\usepackage{enumitem}

\usepackage{booktabs}

\aclfinalcopy 

\title{Adversarial Learning of Privacy-Preserving Text Representations \\ for De-Identification of Medical Records}

\author{Max Friedrich\textsuperscript{1}, Arne Köhn\textsuperscript{2}, Gregor Wiedemann\textsuperscript{1}, Chris Biemann\textsuperscript{1}\\[3mm]
\textsuperscript{1}Language Technology Group, Universität Hamburg\\
\texttt{\{\href{mailto:2mfriedr@informatik.uni-hamburg.de}{2mfriedr},\href{mailto:gwiedemann@informatik.uni-hamburg.de}{gwiedemann},\href{mailto:biemann@informatik.uni-hamburg.de}{biemann}\}@informatik.uni-hamburg.de}\\[3mm]
\textsuperscript{2}Department of Language Science and Technology, Saarland University\\
\texttt{\href{mailto:koehn@coli.uni-saarland.de}{koehn@coli.uni-saarland.de}}}

\begin{document}
\maketitle


\begin{abstract}
De-identification is the task of detecting \ac{phi} in medical text.
It is a critical step in sanitizing \acp{ehr} to be shared for research.
Automatic de-identification classifiers can significantly speed up the sanitization process.
However, obtaining a large and diverse dataset to train such a classifier that works well across many types of medical text poses a challenge as privacy laws prohibit the sharing of raw medical records.
We introduce a method to create privacy-preserving shareable representations of medical text (i.e.\ they contain no \ac{phi}) that does not require expensive manual pseudonymization. 
These representations can be shared between organizations to create unified datasets for training de-identification models.
Our representation allows training a simple \acs{lstm}-\acs{crf} de-identification model to an \fone score of $97.4\%$, which is comparable to a strong baseline that exposes private information in its representation.
A robust, widely available de-identification classifier based on our representation could potentially enable studies for which de-identification would otherwise be too costly.
\end{abstract}

\acresetall

\section{Introduction}\label{sec:introduction}

\begin{figure*}
    \centering
    \begin{tikzpicture}[node distance=1.4cm,font=\small]
\tikzset{every node/.style={inner sep=1mm, outer sep=0mm, line width=0mm}}
\tikzstyle{pre}=[<-,semithick, >=latex]
\tikzstyle{post}=[->, semithick, >=latex]

\tikzset{%
    cascaded/.style = {%
        general shadow = {%
            shadow scale = 1,
            shadow xshift = 1ex,
            shadow yshift = 1ex,
            draw,
            thick,
            fill = white},
        general shadow = {%
            shadow scale = 1,
            shadow xshift = .5ex,
            shadow yshift = .5ex,
            draw,
            thick,
            fill = white},
        fill = white, 
        draw,
        semithick,
        minimum height = 1cm
}}

\node (raw) [cascaded,label=below:Raw patient notes, text width=3cm] {James was admitted to St.\ Thomas\dots};
\node (deid) [cascaded,label=below:{PHI-labeled patient notes}, right=2.2cm of raw, text width=3.5cm] {[James]\textsubscript{Patient} was admitted to [St.\ Thomas]\textsubscript{Hosp}\dots};

\node (pseudo) [cascaded,label=below:Pseudonymized patient notes \faLock, above right=0.3cm and 2.5cm of deid, text width=3.5cm] {[Henry]\textsubscript{Patient} was admitted to [River Clinic]\textsubscript{Hosp}\dots};

\node (repr) [cascaded,label={[align=center]below:Private vector representation\\ of patient notes \faLock}, below right=0.3cm and 2.5cm of deid, text width=3.5cm] {[$\square\square\square$]\textsubscript{Patient} $\square\square\square$ $\square\square\square$ $\square\square\square$ [$\square\square\square$ $\square\square\square$]\textsubscript{Hosp}\dots};

\path[post,every node/.style={sloped,anchor=north,auto=false,align=center}]
(raw) edge node {\scriptsize \itshape PHI labeling / \\[-1mm] \itshape \scriptsize de-identification} (deid)            
(deid.east) edge node {\scriptsize \itshape Pseudonymization} (pseudo.west)
(deid.east) edge node {\scriptsize \itshape Non-reversible \\[-1mm] \itshape \scriptsize transformation} (repr.west);

\path[every node/.style={sloped,anchor=south,auto=false,align=center}]
(raw) edge node {\faUsers} (deid)            
(deid.east) edge node {\faUsers\; \faDollar} (pseudo.west)
(deid.east) edge node {\faCogs} (repr.west);

\node [fit={(raw)([shift={(2mm,2mm)}]pseudo.north east)}, rectangle] {}; 

\end{tikzpicture}
    \caption{Sharing training data for de-identification.
        \ac{phi} annotations are marked with [brackets].
        Upper alternative: traditional process using manual pseudonymization.
        Lower alternative: our approach of sharing private vector representations.
        The people icon represents tasks done by humans; the gears icon represents tasks done by machines; the lock icon represents privacy-preserving artifacts.
        Manual pseudonymization is marked with a dollar icon to emphasize its high costs.
    }\label{fig:process}
\end{figure*}
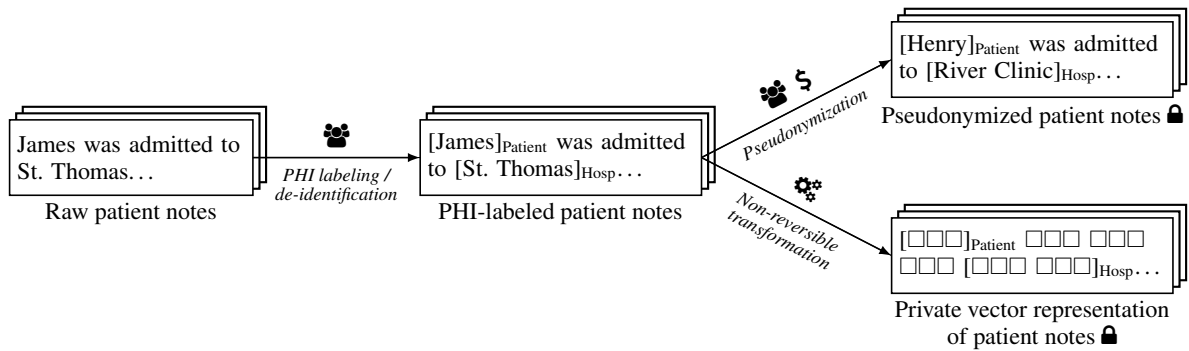

\Acp{ehr} are are valuable resource that could potentially be used in large-scale medical research \citep{botsis2010secondary, birkhead2015uses, cowie2017electronic}.
In addition to structured medical data, \acp{ehr} contain free-text patient notes that are a rich source of information \citep{jensen2012mining}.
However, due to privacy and data protection laws, medical records can only be shared and used for research if they are sanitized to not include information potentially identifying patients.
The \ac{phi} that may not be shared includes potentially identifying information such as names, geographic identifiers, dates, and account numbers; the American \acl{hipaa}\footnote{\url{https://legislink.org/us/pl-104-191}} (\acs{hipaa}, \citeyear{usa1996hipaa}) defines 18 categories of \ac{phi}.
De-identification is the task of finding and labeling \ac{phi} in medical text as a step toward sanitization.
As the information to be removed is very sensitive, sanitization always requires final human verification.
Automatic de-identification labeling can however significantly speed up the process, as shown for other annotation tasks in e.g. \citet{yimam2015narrowing}.

Trying to create an automatic classifier for de-identification leads to a ``chicken and egg problem''~\citep{uzuner2007evaluating}: without a comprehensive training set, an automatic de-identification classifier cannot be developed, but without access to automatic de-identification, it is difficult to share large corpora of medical text in a privacy-preserving way for research (including for training the classifier itself).
The standard method of data protection compliant sharing of training data for a de-identification classifier requires humans to pseudonymize protected information with substitutes in a document-coherent way.
This includes replacing e.g.\ every person or place name with a different name, offsetting dates by a random amount while retaining date intervals, and replacing misspellings with similar misspellings of the pseudonym \cite{uzuner2007evaluating}.

In 2019, a pseudonymized dataset for de-identification from a single source, the \emph{i2b2~2014} dataset, is publicly available \citep{stubbs2015annotating}.
However, de-identification classifiers trained on this dataset do not generalize well to data from other sources~\citep{stubbs2017identification}.
To obtain a universal de-identification classifier, many medical institutions would have to pool their data. But,
preparing this data for sharing using the document-coherent pseudonymization approach requires large human effort \citep{dernoncourt2017identification}.

To address this problem, we introduce an adversarially learned representation of medical text that allows privacy-preserving sharing of training data for a de-identification classifier by transforming text non-reversibly into a vector space and only sharing this representation.
Our approach still requires humans to annotate \ac{phi} (as this is the training data for the actual de-identification task) but the pseudonymization step (replacing \ac{phi} with coherent substitutes) is replaced by the automatic transformation to the vector representation instead.
A classifier then trained on our representation cannot contain any protected data, as it is never trained on raw text (as long as the representation does not allow for the reconstruction of sensitive information).
The traditional approach to sharing training data is conceptually compared to our approach in \cref{fig:process}.

\section{Related Work}\label{sec:related}
Our work builds upon two lines of research: firstly de-identification, as the system has to provide good de-identification performance, and secondly adversarial representation learning, to remove all identifying information from the representations to be distributed.

\subsection{Automatic De-Identification}
Analogously to many \ac{nlp} tasks, the state of the art in de-identification changed in recent years from rule-based systems and shallow machine learning approaches like \acp{crf} \citep{uzuner2007evaluating,meystre2010automatic} to deep learning methods \citep{stubbs2017identification, dernoncourt2017identification, liu2017identification}.

Three i2b2 shared tasks on de-identification were run in 2006 \citep{uzuner2007evaluating}, 2014 \citep{stubbs2015automated}, and 2016 \citep{stubbs2017identification}.
The organizers performed manual pseudonymization on clinical records from a single source to create the datasets for each of the tasks.
An \fone score of $95\%$ has been suggested as a target for reasonable de-identification systems \citep{stubbs2015automated}.

\citet{dernoncourt2017identification} first applied \iac{lstm} \citep{hochreiter1997long} model with \iac{crf} output component to de-identification.
Transfer learning from a larger dataset slightly improves performance on the i2b2 2014 dataset \citep{lee2018transfer}.
\citet{liu2017identification} achieve state-of-the-art performance in de-identification by combining a deep learning ensemble with a rule component.

Up to and including the 2014 shared task, the organizers emphasized that it is unclear if a system trained on the provided datasets will generalize to medical records from other sources \citep{uzuner2007evaluating,stubbs2015automated}.
The 2016 shared task featured a sight-unseen track in which de-identification systems were evaluated on records from a new data source.
The best system achieved an \fone score of $79\%$, suggesting that systems at the time were not able to deliver sufficient performance on completely new data \citep{stubbs2017identification}.

\subsection{Adversarial Representation Learning}
\label{sec:advers-repr-learn}
Fair representations \citep{zemel2013learning,hamm2015preserving} aim to encode features of raw data that allows it to be used in e.g.\ machine learning algorithms while obfuscating membership in a protected group or other sensitive attributes.
The \ac{dann} architecture \citep{ganin2016domain} is a deep learning implementation of a three-party game between a representer, a classifier, and an adversary component.
The classifier and the adversary are deep learning models with shared initial layers.
A gradient reversal layer is used to worsen the representation for the adversary during back-propagation: when training the adversary, the adversary-specific part of the network is optimized for the adversarial task but the shared part is updated against the gradient to make the shared representation less suitable for the adversary.

Although initially conceived for use in domain adaptation, \acp{dann} and similar adversarial deep learning models have recently been used to obfuscate demographic attributes from text \citep{elazar2018adversarial,li2018towards} and subject identity \citep{feutry2018learning} from images.
\citet{elazar2018adversarial} warn that when a representation is learned using gradient reversal methods, continued adversary training on the frozen representation may allow adversaries to break representation privacy.
To test whether the unwanted information is not extractable from the generated information anymore, adversary training needs to continue on the frozen representation after finishing training the system. Only if after continued adversary training the information cannot be recovered, we have evidence that it really is not contained in the representation anymore.

\section{Dataset and De-Identification Model}\label{sec:deidentification-model}

We evaluate our approaches using the i2b2 2014
dataset~\citep{stubbs2015annotating}, which was released as part of
the 2014 i2b2/UTHealth shared task track 1 and is the largest publicly
available dataset for de-identification today.
It contains 1304 free-text documents with \ac{phi} annotations.
The i2b2 dataset uses the 18 categories of \ac{phi} defined by \ac{hipaa} as a starting point for its own set of \ac{phi} categories.
In addition to the \ac{hipaa} set of categories, it includes (sub-)categories such as doctor names, professions, states, countries, and ages under 90.

We compare three different approaches: a non-private de-identification classifier and two privacy-enabled extensions, automatic pseudonymization (\cref{sec:automatic-pseudonymization}) and adversarially learned representations (\cref{sec:adversarial-representation}).

Our non-private system as well as the privacy-enabled extensions are based on a bidirectional \ac{lstm}-\ac{crf} architecture that has been proven to work well in sequence tagging \citep{huang2015bidirectional,lample2016neural} and de-identification \citep{dernoncourt2017identification,liu2017identification}.
We only use pre-trained FastText \citep{bojanowski2017enriching} or GloVe \citep{pennington2014glove} word embeddings, not explicit character embeddings, as we suspect that these may allow easy re-identification of private information if used in shared representations.
In place of learned character features, we provide the casing feature from \citet{reimers2017optimal} as an additional input.
The feature maps words to a one-hot representation of their casing (\textit{numeric}, \textit{mainly numeric}, \textit{all lower}, \textit{all upper}, \textit{initial upper}, \textit{contains digit}, or \textit{other}).

\Cref{tab:deid-hyperparameters} shows our raw de-identification model's hyperparameter configuration that was determined through a random hyperparameter search.

\begin{table}
    \centering
    \begin{tabular}{ll}
     \toprule
     Hyperparameter & Value\\
     \midrule
     Pre-trained embeddings & FastText, GloVe\\
     Casing feature & Yes\\
     Batch size & 32\\
     Number of LSTM layers & 2\\
     LSTM units per layer/dir. & 128\\
     Input embedding dropout & $0.1$\\
     Variational dropout & $0.25$\\
     Dropout after LSTM & $0.5$\\
     Optimizer & Nadam\\
     Gradient norm clipping & $1.0$\\
     \bottomrule
    \end{tabular}
    \caption{Hyperparameter configuration of our de-identification model.}\label{tab:deid-hyperparameters}
\end{table}

\section{Automatic Pseudonymization}\label{sec:automatic-pseudonymization}
To provide a baseline to compare our primary approach against, we introduce a naïve word-level automatic pseudonymization approach that exploits the fact that state-of-the-art de-identification models \citep{liu2017identification,dernoncourt2017identification} as well as our non-private de-identification model work on the sentence level and do not rely on document coherency.
Before training, we shuffle the training sentences and replace all \ac{phi} tokens with a random choice of a fixed number $N$ of their closest neighbors in an embedding space (including the token itself), as determined by cosine distance in a pre-computed embedding matrix.

Using this approach, the sentence
\begin{quote}
    [James] was admitted to [St. Thomas]
\end{quote}
may be replaced by
\begin{quote}
    [Henry] was admitted to [Croix Scott].
\end{quote}
While the resulting sentences do not necessarily make sense to a reader (e.g.\ ``Croix Scott'' is not a realistic hospital name), its embedding representation is similar to the original.
We train our de-identification model on the transformed data and test it on the raw data.
The number of neighbors $N$ controls the privacy properties of the approach: $N = 1$ means no pseudonymization; setting $N$ to the number of rows in a precomputed embedding matrix delivers perfect anonymization but the resulting data may be worthless for training a de-identification model.

\section{Adversarial Representation}\label{sec:adversarial-representation}

\begin{figure}
    \centering
    \begin{tikzpicture}[node distance=1.4cm,font=\small]
\tikzset{every node/.style={inner sep=1mm, outer sep=0mm, line width=0mm}}

\tikzstyle{token}=[rectangle,draw=black,fill=white,semithick,text width=1.1cm, minimum width=1.3cm, minimum height=5mm, text height=1.5ex,text depth=.25ex]
\tikzstyle{model}=[rounded rectangle,draw=black,fill=white,semithick, minimum width=3cm, minimum height=8mm, text height=1.5ex,text depth=.25ex, inner sep=1.5mm]
\tikzstyle{dots} = []
\tikzstyle{vector} = [draw, shape=rectangle, fill=white, semithick, minimum width=1.3cm, minimum height=3.25mm]
\tikzstyle{half vector} = [shape=rectangle, semithick, minimum width=1.3cm, minimum height=3.25mm] 
\tikzstyle{two thirds vector} = [draw, shape=rectangle, fill=white, semithick, minimum width=0.975cm, minimum height=3.25mm]
\tikzstyle{quarter vector} = [draw, shape=rectangle, fill=white, semithick, minimum width=3.25mm, minimum height=3.25mm]
\tikzstyle{pre}=[<-,semithick, >=latex]
\tikzstyle{post}=[->, semithick, >=latex]

\node[token] (mr input){James};
\node[token, right of=mr input] (smith input) {was};
\node[token, right of=smith input] (was input) {admitted};
\node[dots, right=1mm of was input] (input dots) {$\cdots$};
    
\node[vector, above=8mm of mr input] (mr embedding) {};
\node[vector, right of=mr embedding] (smith embedding) {};
\node[vector, right of=smith embedding] (was embedding) {};
\node[dots, right=1mm of was embedding] (embedding dots) {$\cdots$};

\begin{scope}[on background layer]
    \node (input box) [draw,fill=black!5,fit=(mr input) (input dots), inner sep=1.5mm] {};
    \node (feature box) [draw,fill=black!5,fit=(mr embedding) (embedding dots), inner sep=1.5mm] {};
\end{scope}

\node[model,above=5mm of feature box] (representation model) {Representation Model};

\node[vector, above=2.2cm of mr embedding] (mr representation) {};
\node[vector, right of=mr representation] (smith representation) {};
\node[vector, right of=smith representation] (was representation) {};
\node[dots, right=1mm of was representation] (representation dots) {$\cdots$};

\begin{scope}[on background layer]
    \node (representation box) [draw,fill=black!5,fit=(mr representation) (representation dots), inner sep=1.5mm] {};
\end{scope}

\node[model,above left=8mm and -2cm of representation box] (deid model) {De-Identification Model};

\node[two thirds vector, above left=2.5cm and 0.5cm of mr representation] (mr output) {};
\node[two thirds vector, right=3mm of mr output] (smith output) {};
\node[two thirds vector, right= 3mm of smith output] (was output) {};
\node[dots, right=3mm of was output] (output dots) {$\cdots$};    

\begin{scope}[on background layer]
 \node (output box) [draw,fill=black!5,fit=(mr output) (output dots), inner sep=1.5mm] {};
\end{scope}

\node[model,above right=8mm and -1.5cm of representation box] (adversary model) {Adversary Model};

\node[quarter vector, above right=2.5cm and 4mm of was representation] (adversary output) {};

\begin{scope}[on background layer]
    \node (adversary output box) [draw,fill=black!5,fit=(adversary output), inner sep=1.5mm] {};
\end{scope}

\foreach \i in {mr embedding, smith embedding, was embedding, mr representation, smith representation, was representation} {
    \draw[semithick] (\i.north west) rectangle ($(\i.north west) + (3.25mm, -3.25mm)$);
    \draw[semithick] (\i.north west) rectangle ($(\i.north west) + (6.5mm, -3.25mm)$);
    \draw[semithick] (\i.north west) rectangle ($(\i.north west) + (9.75mm, -3.25mm)$);    
};

\foreach \i in {mr output, smith output, was output} {
    \draw[semithick] (\i.north west) rectangle ($(\i.north west) + (3.25mm, -3.25mm)$);
    \draw[semithick] (\i.north west) rectangle ($(\i.north west) + (6.5mm, -3.25mm)$);
};

\path[post] (input box) edge (feature box);
\path[post] (feature box) edge (representation model);
\path[post] (representation model) edge (representation box);

\path[post] (representation box) edge (deid model);
\path[post] (deid model) edge (output box);

\path[post] (representation box) edge (adversary model);
\path[post,dotted] (feature box.east) edge[bend right=30] (adversary model);
\path[post,dotted] (representation box.north east) edge (adversary model);
\path[post] (adversary model) edge (adversary output box);

\node[anchor=east, left=1mm of input box] {Tokens};
\node[anchor=east, left=1mm of feature box] {Emb.};
\node[anchor=east, left=1mm of representation box] {Represent.};
\node[anchor=center, above=1mm of output box] {De-identification output};
\node[anchor=center, above=1mm of adversary output box] {Adversary output};

\end{tikzpicture}
    \caption[Adversarial model architecture]{%
        Simplified visualization of the adversarial model architecture.
        Sequences of squares denote real-valued vectors, dotted arrows represent possible additional real or fake inputs to the adversary.
        The casing feature that is provided as a second input to the de-identification model is omitted for legibility.}\label{fig:adversarial-model}
\end{figure}
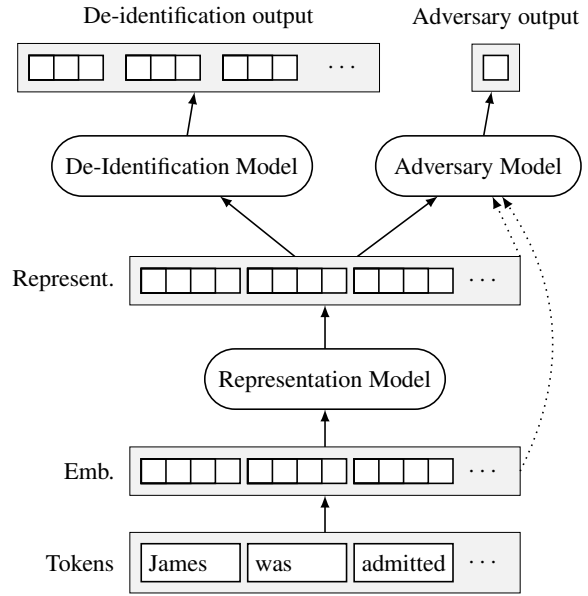

We introduce a new data sharing approach that is based on an adversarially learned private representation and improves on the pseudonymization from \cref{sec:automatic-pseudonymization}.
After training the representation on an initial publicly available dataset, e.g.\ the i2b2 2014 data, a central model provider shares the frozen representation model with participating medical institutions.
They transform their \ac{phi}-labeled raw data into the pseudonymized representation, which is then pooled into a new public dataset for de-identification.
Periodically, the pipeline consisting of the representation model and a trained de-identification model can be published to be used by medical institutions on their unlabeled data.

Since both the representation model and the resulting representations are shared in this scenario, our representation procedure is required to prevent two attacks:
\begin{enumerate}[label=A\arabic*.,ref=A\arabic*]
    \item Learning an inverse representation model that transforms representations back to original sentences containing \ac{phi}.\label{item:attack1}
    \item Building a lookup table of inputs and their exact representations that can be used in known plaintext attacks.\label{item:attack2}
\end{enumerate}

\subsection{Architecture}
Our approach uses a model that is composed of three components: a representation model, the de-identification model from \cref{sec:deidentification-model}, and an adversary.
An overview of the architecture is shown in \cref{fig:adversarial-model}.

The representation model maps a sequence of word embeddings to an intermediate vector representation sequence.
The de-identification model receives this representation sequence as an input instead of the original embedding sequence.
It retains the casing feature as an auxiliary input.
The adversary has two inputs, the representation sequence and an additional embedding or representation sequence, and a single output unit.

\subsection{Representation}
To protect against \ref{item:attack1}, our representation must be invariant to small input changes, like a single \ac{phi} token being replaced with a neighbor in the embedding space.
Again, the number of neighbors $N$ controls the privacy level of the representation.

To protect against \ref{item:attack2}, we add a random element to the representation that makes repeated transformations of one sentence indistinguishable from representations of similar input sentences.

We use a bidirectional \ac{lstm} model to implement the representation.
It applies Gaussian noise $\bm{N}$ with zero mean and trainable standard deviations to the input embeddings $\bm{E}$ and the output sequence.
The model learns a standard deviation for each of the input and output dimensions.

\begin{align}
\bm{R} = \bm{N}_{\text{out}} + \text{LSTM}(\bm{E} + \bm{N}_{\text{in}})
\end{align}

In a preliminary experiment, we confirmed that adding noise with a single, fixed standard deviation is not a viable approach for privacy-preserving representations.
To change the cosine similarity neighborhoods of embeddings at all, we need to add high amounts of noise (more than double of the respective embedding matrix's standard deviation), which in turn results in \emph{unrealistic} embeddings that do not allow training a de-identification model of sufficient quality.
In contrast to the automatic pseudonymization approach from \cref{sec:automatic-pseudonymization} that only perturbs \ac{phi} tokens, the representation models in this approach processes all tokens to represent them in a new embedding space.
We evaluate the representation sizes $d \in \{50, 100, 300\}$. 

\subsection{Adversaries}
We use two adversaries that are trained on tasks that directly follow from \ref{item:attack1} and \ref{item:attack2}:
\begin{enumerate}[label=T\arabic*.,ref=T\arabic*]
    \item Given a representation sequence and an embedding sequence, decide if they were obtained from the same sentence.
    \item Given two representation sequences (and their cosine similarities), decide if they were obtained from the same sentence.
\end{enumerate}
We generate the representation sequences for the second adversary from a copy of the representation model with shared weights.
We generate real and fake pairs for adversarial training using the automatic pseudonymization approach presented in \cref{sec:automatic-pseudonymization}, limiting the number of replaced \ac{phi} tokens to one per sentence.

The adversaries are implemented as bidirectional \ac{lstm} models with single output units.
We confirmed that this type of model is able to learn the adversarial tasks on random data and raw word embeddings in preliminary experiments.
To use the two adversaries in our architecture, we average their outputs.

 \begin{figure}
    \centering
    \begin{tikzpicture}[node distance=1.5cm,font=\scriptsize, sibling distance=6.5mm, level distance=1cm, grow=up, edge from parent/.style = {->, >=latex, draw}]
\tikzset{every node/.style={inner sep=0mm, outer sep=0mm, line width=0mm}}

\tikzstyle{model}=[rounded rectangle,draw=black,fill=white, minimum width=5.5mm, minimum height=3mm, semithick, text depth=-.2ex]
\tikzstyle{train}=[ultra thick]
\tikzstyle{label}=[text height=0.75ex,text depth=0ex]
\tikzstyle{dots} = []
\tikzstyle{pre}=[<-,semithick, >=latex]
\tikzstyle{post}=[->, semithick, >=latex]

\node[model, train] (a) {R}
    child {node[model] (a adv) {A}}
    child {node[model, train] (a deid) {D}};

\node[label, below= 2mm of a] (a label) {1.}; 

\node[model, right=of a] (b) {R}
    child {node[model, train] (b adv) {A}}
    child {node[model] (b deid) {D}};

\node[label, below=2mm of b] (b label) {2.}; 

\node[model, right=of b] (c) {R}
    child {node[model, train] (c adv) {A}}
    child {node[model, train] (c deid) {D}};
    
\node[label, below=2mm of c] (c label) {3.$\,$a)}; 

\node[model, right= of c, train] (d) {R}
    child {node[model] (d adv) {A}}
    child {node[model] (d deid) {D}};
    
\node[label, below=2mm of d] (d label) {3.$\,$b)}; 
    
\begin{scope}[on background layer]
    \node (a box) [draw,fill=black!5,fit=(a) (a label) (a adv) (a deid), inner sep=1mm] {};
    \node (b box) [draw,fill=black!5,fit=(b) (b label) (b adv) (b deid), inner sep=1mm] {};
    \node (c box) [draw,fill=black!5,fit=(c) (c label) (c adv) (c deid), inner sep=1mm] {};
    \node (d box) [draw,fill=black!5,fit=(d) (d label) (d adv) (d deid), inner sep=1mm] {};
\end{scope}

\path[post] (a box) edge (b box);
\path[post] (b box) edge (c box);
\path[post] (c box) edge[bend right=10] (d box);
\path[post] (d box) edge[bend right=10] (c box);

\end{tikzpicture}
    \caption[Adversarial training procedure]{%
        Visualization of \citeauthor{feutry2018learning}'s three-part training procedure.
        The adversarial model layout follows \cref{fig:adversarial-model}: the representation model is at the bottom, the left branch is the de-identification model and the right branch is the adversary.
        In each step, the thick components are trained while the thin components are frozen.
    }\label{fig:feutry-training}
\end{figure}
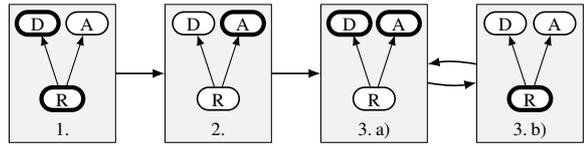

\subsection{Training}
We evaluate two training procedures: \ac{dann} training~\citep{ganin2016domain} and the three-part procedure from \citet{feutry2018learning}.

In \ac{dann} training, the three components are trained conjointly, optimizing the sum of losses.
Training the de-identification model modifies the representation model weights to generate a more meaningful representation for de-identification.
The adversary gradient is reversed with a gradient reversal layer between the adversary and the representation model in the backward pass, causing the representation to become less meaningful for the adversary.

The training procedure by \citet{feutry2018learning} is shown in \cref{fig:feutry-training}.
It is composed of three phases:
\begin{enumerate}[label=P\arabic*.,ref=P\arabic*]
    \item The de-identification and representation models are pre-trained together, optimizing the de-identification loss $l_{\text{deid}}$.
    \item The representation model is frozen and the adversary is pre-trained, optimizing the adversarial loss $l_{\text{adv}}$.
    \item In alternation, for one epoch each:
    \begin{enumerate}
        \item The representation is frozen and both de-identification model and adversary are trained, optimizing their respective losses $l_{\text{deid}}$ and $l_{\text{adv}}$.
        \item The de-identification model and adversary are frozen and the representation is trained, optimizing the combined loss
        \begin{align}
            l_{\text{repr}} = l_{\text{deid}} + \lambda \abs{l_{\text{adv}} - l_{\text{random}}}
        \end{align}
        \label{item:repr-training}
    \end{enumerate}
\end{enumerate}
In each of the first two phases, the respective validation loss is monitored to decide at which point the training should move on to the next phase.
The alternating steps in the third phase each last one training epoch; the early stopping time for the third phase is determined using only the combined validation loss from Phase \ref{item:repr-training}.

Gradient reversal is achieved by optimizing the combined representation loss while the adversary weights are frozen.
The combined loss is motivated by the fact that the adversary performance should be the same as a random guessing model, which is a lower bound for anonymization~\citep{feutry2018learning}.
The term $\abs{l_{\text{adv}} - l_{\text{random}}}$ approaches $0$ when the adversary performance approaches random guessing\footnote{In the case of binary classification: $L_{\text{random}} = -\log \frac{1}{2}$.}.
$\lambda$ is a weighting factor for the two losses; we select $\lambda=1$.

\section{Experiments}
To evaluate our approaches, we perform experiments using the i2b2 2014 dataset.

\paragraph{Preprocessing:}
We apply aggressive tokenization similarly to \citet{liu2017identification}, including splitting at all punctuation marks and mid-word e.g.\ if a number is followed by a word (``25yo'' is split into ``25'', ``yo'') in order to minimize the amount of GloVe out-of-vocabulary tokens.
We extend spaCy's\footnote{\url{https://spacy.io}} sentence splitting heuristics with additional rules for splitting at multiple blank lines as well as bulleted and numbered list items.

\paragraph{Deep Learning Models:}
We use the Keras framework\footnote{\url{https://keras.io}} \citep{chollet2015keras} with the TensorFlow backend \citep{abadi2015tensorflow} to implement our deep learning models.

\paragraph{Evaluation:}
In order to compare our results to the state of the art, we use the token-based binary \ac{hipaa} \fone score as our main metric for de-identification performance.
\citet{dernoncourt2017identification} deem it the most important metric: deciding if an entity is \ac{phi} or not is generally more important than assigning the correct category of \ac{phi}, and only \ac{hipaa} categories of \ac{phi} are required to be removed by American law.
Non-\ac{phi} tokens are not incorporated in the \fone score.
We perform the evaluation with the official i2b2 evaluation script\footnote{\url{https://github.com/kotfic/i2b2\_evaluation\_scripts}}.

\section{Results}
\Cref{tab:baseline-results} shows de-identification performance results for the non-private de-identification classifier (upper part, in comparison to the state of the art) as well as the two privacy-enabled extensions (lower part).
The results are average values out of five experiment runs.

\subsection{Non-private De-Identification Model}
When trained on the raw i2b2 2014 data, our models achieve \fone scores that are comparable to \citeauthor{dernoncourt2017identification}'s results.
The casing feature improves GloVe by $0.4$ percentage points.

\begin{table}
    \centering
    \begin{tabular}{ll}
        \toprule
         Model & \fone (\%)\\
        \midrule 
         Our non-private FastText & $97.67$ \\
         Our non-private GloVe & $97.24$ \\
         Our non-private GloVe + casing & $97.62$ \\
         \addlinespace
         \citeauthor{dernoncourt2017identification} (\ac{lstm}-\ac{crf}) & $97.85$\\
         \citeauthor{liu2017identification} (ensemble + rules) & $\bm{98.27}$\\
         \midrule
         Our autom. pseudon. FastText & $96.75$\\
         Our autom. pseudon. GloVe & $96.42$\\
         \addlinespace
         Our adv. repr. FastText & $\bm{97.40}$\\
         Our adv. repr. GloVe & $96.89$\\
         \bottomrule
    \end{tabular}
    \caption{Binary \ac{hipaa} \fone scores of our non-private (top) and private (bottom) de-identification approaches on the i2b2 2014 test set in comparison to non-private the state of the art. Our private approaches use $N=100$ neighbors as a privacy criterion.}\label{tab:baseline-results}
\end{table}

\subsection{Automatic Pseudonymization}
\begin{figure}
    \centering
    \input{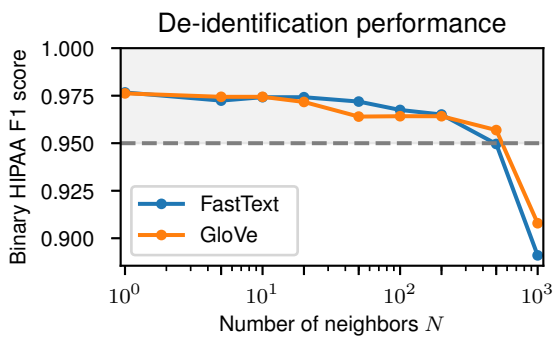}
    \caption[De-identification with automatic pseudonymization]{%
        \fone scores of our models when trained on automatically pseudonymized data where \ac{phi} tokens are moved to one of different numbers of neighbors $N$.
        The gray dashed line marks the $95\%$ target \fone score.
}\label{fig:auto-pseudo}
\end{figure}

For both FastText and GloVe, moving training \ac{phi} tokens to random tokens from up to their $N=200$ closest neighbors does not significantly reduce de-identification performance (see \cref{fig:auto-pseudo}).
\fone scores for both models drop to around $95\%$ when selecting from $N=500$ neighbors and to around $90\%$ when using $N=1\,000$ neighbors.
With $N=100$, the FastText model achieves an \fone score of $96.75\%$ and the GloVe model achieves an \fone score of $96.42\%$.

\subsection{Adversarial Representation}
We do not achieve satisfactory results with the conjoint \ac{dann} training procedure: in all cases, our models learn representations that are not sufficiently resistant to the adversary.
When training the adversary on the frozen representation for an additional $20$ epochs, it is able to distinguish real from fake input pairs on a test set with accuracies above $80\%$.
This confirms the difficulties of \ac{dann} training as described by \citet{elazar2018adversarial} (see Section~\ref{sec:advers-repr-learn}).

In contrast, with the three-part training procedure, we are able to learn a representation that allows training a de-identification model while preventing an adversary from learning the adversarial tasks, even with continued training on a frozen representation.

\begin{figure*}
    \centering
    \input{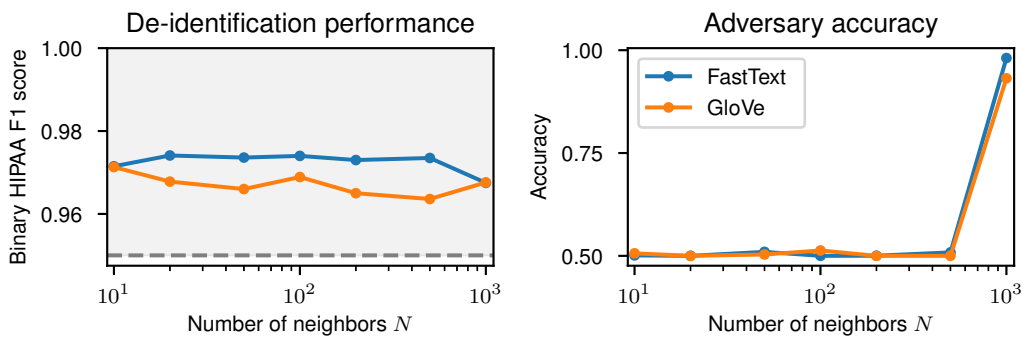}
    \caption[De-identification with adversarially learned representations]{%
        Left: de-identification \fone scores of our models using an adversarially trained representation with different numbers of neighbors $N$ for the representation invariance requirement.
        Right: mean adversary accuracy when trained on the frozen representation for an additional $50$ epochs.
        The figure shows average results out of five experiment runs.
    }\label{fig:adversarial-deid}
\end{figure*}

\Cref{fig:adversarial-deid} (left) shows our de-identification results when using adversarially learned representations.
A higher number of neighbors $N$ means a stronger invariance requirement for the representation.
For values of $N$ up to $1\,000$, our FastText and GloVe models are able to learn representations that allow training de-identification models that reach or exceed the target \fone score of $95\%$.
However, training becomes unstable for $N>500$: at this point, the adversary is able to break the representation privacy when trained for an additional $50$ epochs (\cref{fig:adversarial-deid} right).

Our choice of representation size $d \in \{50, 100, 300\}$ does not influence de-identifi\-ca\-tion or adversary performance, so we select $d=50$ for further evaluation.
For $d=50$ and $N=100$, the FastText model reaches an \fone score of $97.4\%$ and the GloVe model reaches an \fone score of $96.89\%$.

\section{De-Identification Performance}

In the following, we discuss the results of our models with regard to our goal of sharing sensitive training data for automatic de-identification.
Overall, privacy-preserving representations come at a cost, as our best privacy-preserving model scores $0.27$ points \fone score lower than our best non-private model; we consider this relative increase of errors of less than $10\%$ as tolerable.

\paragraph{Raw Text De-Identification:}
We find that the choice of GloVe or FastText embeddings does not meaningfully influence de-identification performance.
FastText's approach to embedding unknown words (word embeddings are the sum of their subword embeddings) should intuitively prove useful on datasets with misspellings and ungrammatical text.
However, when using the additional casing feature, FastText beats GloVe only by $0.05$ percentage points on the i2b2 test set.
In this task, the casing feature makes up for GloVe's inability to embed unknown words.

\citet{liu2017identification} use a deep learning ensemble in combination with hand-crafted rules to achieve state-of-the-art results for de-identification.
Our model's scores are similar to the previous state of the art, a bidirectional \ac{lstm}-\ac{crf} model with character features \citep{dernoncourt2017identification}.

\paragraph{Automatically Pseudonymized Data:}
Our naïve automatic word-level pseudonymization approach allows training reasonable de-iden\-tification models when selecting from up to $N=500$ neighbors.
There is almost no decrease in \fone score for up to $N = 20$ neighbors for both the FastText and GloVe model.

\paragraph{Adversarially Learned Representation:}
Our adversarially trained vector representation allows training reasonable de-identification models (\fone scores above $95\%$) when using up to $N=1\,000$ neighbors as an invariance requirement.
The adversarial representation results beat the automatic pseudonymization results because the representation model can act as a task-specific feature extractor.
Additionally, the representations are more general as they are invariant to word changes.

\section{Privacy Properties}

In this section, we discuss our models with respect to their privacy-preserving properties.

\paragraph{Embeddings:}
When looking up embedding space neighbors for words, it is notable that many FastText neighbors include the original word or parts of it as a subword.
For tokens that occur as \ac{phi} in the i2b2 training set, on average $7.37$ of their $N=100$ closest neighbors in the FastText embedding matrix contain the original token as a subword.
When looking up neighbors using GloVe embeddings, the value is $0.44$.
This may indicate that FastText requires stronger perturbation (i.e.\ higher $N$) than GloVe to sufficiently obfuscate protected information.

\paragraph{Automatically Pseudonymized Data:}
The word-level pseudonymization does not guarantee a minimum perturbation for every word, e.g.\ in a set of pseudonymized sentences using $N = 100$ FastText neighbors, we found the phrase
\begin{quote}
    [Florida Hospital],
\end{quote}
which was replaced with
\begin{quote}
    [Miami-Florida Hosp].
\end{quote}

Additionally, the approach may allow an adversary to piece together documents from the shuffled sentences.
If multiple sentences contain similar pseudonymized identifiers, they will likely come from the same original document, undoing the privacy gain from shuffling training sentences across documents.
It may be possible to infer the original information using the overlapping neighbor spaces.
To counter this, we can re-introduce document-level pseudonymization, i.e.\ moving all occurrences of \iac{phi} token to the same neighbor.
However, we would then also need to detect misspelled names as well as other hints to the actual tokens and transform them similarly to the original, which would add back much of the complexity of manual pseudonymization that we try to avoid.

\paragraph{Adversarially Learned Representation:}
Our adversarial representation empirically satisfies a strong privacy criterion: representations are invariant to \textit{any} protected information token being replaced with \textit{any} of its $N$ neighbors in an embedding space.
When freezing the representation model from an experiment run using up to $N = 500$ neighbors and training the adversary for an additional $50$ epochs, it still does not achieve higher-than-chance accuracies on the training data.
Due to the additive noise, the adversary does not overfit on its training set but rather fails to identify any structure in the data.

In the case of $N = 1\,000$ neighbors, the representation never becomes stable in the alternating training phase.
The adversary is always able to break the representation privacy.

\section{Conclusions \& Future Work}
We introduced a new approach to sharing training data for de-identification that requires
lower human effort than the existing approach of document-coherent pseudonymization.
Our approach is based on adversarial learning, which yields representations that can be distributed since they do not contain private health information.
The setup is motivated by the need of de-identification of medical text before sharing; our approach provides a lower-cost alternative than manual pseudonymization and gives rise to the pooling of de-identification datasets from heterogeneous sources in order to train more robust classifiers.
Our implementation and experimental data are publicly available\footnote{\url{https://github.com/maxfriedrich/deid-training-data}}.

As precursors to our adversarial representation approach, we developed a deep learning model for de-identification that does not rely on explicit character features as well as an automatic word-level pseudonymization approach.
A model trained on our automatically pseudonymized data with $N=100$ neighbors loses around one percentage point in \fone score when compared to the non-private system, scoring $96.75\%$ on the i2b2 2014 test set.

Further, we presented an adversarial learning based private representation of medical text that is invariant to any \ac{phi} word being replaced with any of its embedding space neighbors and contains a random element.
The representation allows training a de-identification model while being robust to adversaries trying to re-identify protected information or building a lookup table of representations.
We extended existing adversarial representation learning approaches by using two adversaries that discriminate real from fake sequence pairs with an additional sequence input.

The representation acts as a task-specific feature extractor.
For an invariance criterion of up to $N=500$ neighbors, training is stable and adversaries cannot beat the random guessing accuracy of $50\%$.
Using the adversarially learned representation, de-identification models reach an \fone score of $97.4\%$, which is close to the non-private system (\(97.67\%\)).
In contrast, the automatic pseudonymization approach only reaches an \fone score of \(95.0\%\) at \(N = 500\).

Our adversarial representation approach enables cost-effective private sharing of training data for sequence labeling. Pooling of training data for de-identification from multiple institutions would lead to much more robust classifiers.
Eventually, improved de-identification classifiers could help enable large-scale medical studies that eventually improve public health.

\paragraph{Future Work:} The automatic pseudonymization approach could serve as a data augmentation scheme to be used as a regularizer for de-identification models.
Training a model on a combination of raw and pseudonymized data may result in better test scores on the i2b2 test set, possibly improving the state of the art.

Private character embeddings that are learned from a perturbed source could be an interesting extension to our models.

In adversarial learning with the three-part training procedure, it might be possible to tune the $\lambda$ parameter and define a better stopping condition that avoids the unstable characteristics with high values for $N$ in the invariance criterion.
A further possible extension is a dynamic noise level in the representation model that depends on the \ac{lstm} output instead of being a trained weight.
This might allow using lower amounts of noise for certain inputs while still being robust to the adversary.

When more training data from multiple sources become available in the future, it will be possible to evaluate our adversarially learned representation against unseen data.

\section*{Acknowledgments}
This work was partially supported by BWFG Hamburg within the ``Forum 4.0'' project as part of the ahoi.digital funding line.

De-identified clinical records used in this research were provided by the i2b2 National Center for Biomedical Computing funded by U54LM008748 and were originally prepared for the Shared Tasks for Challenges in NLP for Clinical Data organized by Özlem Uzuner, i2b2 and SUNY.

\bibliography{deidentification-clean}
\bibliographystyle{acl_natbib}

\end{document}